\documentclass[runningheads]{llncs}

\usepackage{graphicx}
\usepackage[hidelinks]{hyperref}
\usepackage{algorithm2e}
\RestyleAlgo{ruled}
\usepackage{enumitem}
\usepackage{amssymb}
\usepackage{amsmath, nccmath}
\usepackage{tabularx}
\usepackage{booktabs}
\usepackage{xcolor}
\usepackage{multicol}
\usepackage{breakurl}

\newcolumntype{C}{>{\centering\arraybackslash}X}
\newcolumntype{R}{>{\raggedleft\arraybackslash}X}

\title{Representation Learning of Multivariate \\ Time Series using Attention and Adversarial Training}

\titlerunning{Representation Learning of Multivariate Time Series}

\author{Leon Scharwächter\inst{1,2} \and
Sebastian Otte\inst{2}}%\orcidID{0000-0002-0305-0463}}
\authorrunning{L. Scharwächter and S. Otte}

\institute{Neuro-Cognitive Modeling Group, Department of Computer Science, University of Tübingen \and Adaptive AI Lab, Institute of Robotics and Cognitive Systems, University of Lübeck}

\def\slen{l} % können wir später noch ändern.

\begin{document}

\maketitle

\begin{abstract}
A critical factor in trustworthy machine learning is to develop robust representations of the training data. Only under this guarantee methods are legitimate to artificially generate data, for example, to counteract imbalanced datasets or provide counterfactual explanations for blackbox decision-making systems. In recent years, Generative Adversarial Networks (GANs) have shown considerable results in forming stable representations and generating realistic data. While many applications focus on generating image data, less effort has been made in generating time series data, especially multivariate signals. In this work, a Transformer-based autoencoder is proposed that is regularized using an adversarial training scheme to generate artificial multivariate time series signals. The representation is evaluated using t-SNE visualizations, Dynamic Time Warping (DTW) and Entropy scores. Our results indicate that the generated signals exhibit higher similarity to an exemplary dataset than using a convolutional network approach.
\end{abstract}
%\begin{multicols}{2}

%\setcounter{footnote}{1}
%\renewcommand*{\thefootnote}{\fnsymbol{footnote}}
%\footnotetext{Corresponding author:\\ %\email{leon.scharwaechter@student.uni-tuebingen.de}}

%\renewcommand*{\thefootnote}{\arabic{footnote}}
%\setcounter{footnote}{0}

\section{Introduction}
Learning patterns in multivariate time series signals has been gaining popularity due an increasing use of real-time sensors and recent advances in deep learning architectures \cite{Assaf2019}. Electrocardiogram records, climate measurements, or motion sensors are some examples of multivariate time series, which typically involve not only temporal dependencies per variable but also dependencies between multiple variables. The analysis of such inherent multivariate temporal patterns led to a wide range of applications like early detection of heart diseases \cite{He2015,Bui2017,Balaha2022}, seismic activity forecasting \cite{MoralesEsteban2010,Moustra2011}, or gesture recognition in human-machine-interaction \cite{Ding2020,Severin2020}. However, deep learning models require large amounts of data to train successfully and available datasets are often scarce or imbalanced. Training deep learning models on small datasets consequently results in overfitting and low generalization capabilities. Moreover, for time series, the possibilities to use data augmentation tricks to artificially expand the datasets are limited due to the need to preserve the inherent structural properties of the examined signals \cite{lang2023generating}. 

Representation learning refers to discovering the most relevant and informative features from the input data, where typically a lower dimensional latent encoding is learned. For this purpose, autoencoders often serve as a basic framework. These are neural network architectures that consist of an encoder and a decoder part, both containing trainable parameters. The encoder transforms the input into a latent encoding, while the decoder creates a reconstruction of the true data given the latent encoding. For example for univariate time series signals, in \cite{Guidotti2020} the authors used a convolutional autoencoder to compress univariate time series signals into a low-dimensional encoding. Afterwards, a local neighborhood is sampled around the latent code of a certain input query. These samples are then decoded to create counterfactual explanations. This is a common procedure in the field of explainable machine learning \cite{Ates2021}. Although using a compressed representation increases the control over the latent features vanilla autoencoders do not guarantee any structure or organization of the latent space and lack of plausibility of the generated samples.

\subsection{Generative Adversarial Networks}
A Generative Adversarial Network (GAN) is a model that is suitable to shape the latent space by enforcing a structure that follows the distribution of the given data \cite{Goodfellow2014}. There have been countless successes using GANs in the domains of computer vision or natural language processing, such as image generation \cite{ledig2017} or text-to-image synthesis \cite{zhang2017}. The utilization of GANs in the field of time series for artificial signal generation and signal forecasting has been gaining traction as well \cite{Brophy2021}. A GAN consists of a generator and a discriminator part, typically initialized as neural networks. The generator $G(z)$ samples a random latent point $z$ from a prior distribution $p_{noise}$ and transforms it into an output of the same dimension of the true data, i.e. generates a fake sample. The discriminator $D(x)$ is a binary classifier that learns to distinguish the fake samples from the true data, i.e. calculates the probability that a data point $x$ is a sample from the data distribution $p_{data}$ rather than a sample from the generative process. $G(z)$ is however optimized to confuse $D(x)$ into believing that the fake samples come from the data distribution. Both networks are optimized simultaneously until they reach an equilibrium, where the solution of this adversarial scheme can be expressed as follows \cite{Goodfellow2014}:
\begin{equation}\label{eq1}
\begin{split}
\min_{G}\max_{D} ~ &\mathbb{E}_{x\sim p_{data}}\log D(x) +\\
&\mathbb{E}_{z\sim p_{noise}}\log(1-D(G(z)))
\end{split}
\end{equation}
The authors furthermore provide a theoretical proof that given enough model capacity and training time, the generator shapes the distribution of the latent space by a mapping $G(z)$ such that $p_{noise} = p_{data}$. For real-world applications it is important to mention that even after this convergence implausible data points can be sampled from the latent space if the dataset consists of noisy or incomplete data.

\subsection{Transformer}
The Transformer is a well-established deep learning architecture that was initially proposed for natural language translation tasks \cite{vaswani2017}. Since then, it has become prevalent in many different domains and has surpassed other models such as convolutional or recurrent neural networks \cite{Khan2022,Lin2022}. Based on an encoder-decoder scheme, it first encodes the input signal into a latent memory using a multi-headed self-attention mechanism over the whole sequence. This enables the Transformer to capture long-term dependencies within the signal, while multiple attention heads can consider different representation structures. The attention mechanism thereby transforms the input into $m$ distinct query, key, and value matrices $Q_{i}, K_{i}$, and $V_{i}$ through trainable, linear projections. Each head $i$ recombines the value matrix $V_{i}$ into the head's output matrix $O_{i}$ via the following scheme (or a similar variant thereof):
\begin{align}
\label{eq2}
%O_m = \operatorname{softmax}\left(\frac{Q_mK_m^T}{\sqrt{d_k}}V_m\right),
O_i = \operatorname{softmax}\left(\frac{Q_iK_i^\top}{\sqrt{d_k}}\right) V_i,
\end{align}
where here $\sqrt{d_k}$ is a normalization constant. The final output of the multi-head attention layer consists of a linear projection of the concatenation $O_1,...,O_m$.

%The decoder then queries the keys of the encoder and uses its values to decode the memory.
%Alternative, aber vllt noch recht raw:
This latent representation is then used by the decoder to generate an output time series. Thereby, for each position in the output, the decoder queries which parts (keys) at the input were most crucial to predict the next Token and uses its values to calculate an output probability. During training, the known output is shifted to the right, because the first position already serves as a label for the prediction. To prevent the decoder of using information that lie in the future, a masking operation is applied; for example, by setting all preceding values in the softmax function to $-\infty$.

\subsection{Problem Formulation}
Inspired by the work of \cite{zhang2022} and \cite{Makhzani2015} an autoencoder is a suitable candidate architecture for stabilizing GAN training. Not only does it contribute in learning a latent representation of a complex distribution, but also reduces mode collapse \cite{Tran2018dist}
%\footnote{Mode collapse describes a problem in GAN training where the generator only learns to produce a single type of output or a small set of outputs}
and makes it easier to perform complex modifications, e.g. through interpolation in the latent space. Since a Transformer comprises an encoder-decoder network, it can easily be adapted into an autoencoder framework that projects the input into a lower dimensionality using a bottleneck. There have been already previous work that utilize GAN training for Transformer networks, for example in developing frameworks for univariate \cite{wu2020} or multivariate \cite{zhang2022} time series forecasting. In \cite{li2022} the authors used a GAN scheme solely based on Transformer encoders for time series representation learning. The goal of this work is to develop an autoencoder for multivariate time series representation learning that is based on both the encoder and decoder of a Transformer network, and organize the latent space by incorporating a GAN training scheme. This way, new plausible time series samples should be generated artificially. The procedure is accompanied by an example dataset. Afterwards the latent space is evaluated using Dynamic Time Warping (DTW), Entropy scores, and t-SNE representations \cite{vandermaaten2008}.

\section{Methods}
\subsection{Model Architectures}
\label{subsec:modelarchitectures}
The model is based on a Transformer network, i.e. an encoder $\zeta_T$ and a decoder $\eta_T$. While the original paper serves as a reference for a detailed description of each individual component \cite{vaswani2017}, here the changes that turn the model into an adversarial autoencoder for multivariate time series data are presented. The input $X \in \mathbb{R}^{\slen \times v}$ is a multivariate time series of length $\slen$ and $v$ numbers of variables and thus comprises a sequence of $\slen$ feature vectors $x_t \in \mathbb{R}^v$ for $x = [x_1, x_2,...,x_\slen]$. As the model operates on continuous data instead on sequences of discrete word indices, the \textit{Embedding Layer} projects the feature vectors $x_t$ into a $d$-dimensional vector space: $u_t = W_ex_t + b_e$, where $d$ is the model dimension, $W_e \in \mathbb{R}^{d \times v}$, $b_e \in \mathbb{R}^d$ are learnable parameters and $u_t \in \mathbb{R}^d : U \in \mathbb{R}^{\slen \times d} = [u_1, u_2,...,u_\slen]$ represents the input that corresponds to the word vectors of the original paper. This method is proposed by \cite{Zerveas2021}. Subsequently, since the Transformer architecture is insensitive to any sequential ordering of the input, a \textit{Positional Encoding Layer} adds a notion of time dependence. This is done by sinusoidal encodings $W_{pos}$ as proposed in the original paper: $U' = U + W_{pos}$, where $W_{pos} \in \mathbb{R}^{\slen \times d}$ contains sines and cosines of different frequencies per model dimension $d$. The Transformer encoder $\zeta_T$ then computes the latent memory $Z \in \mathbb{R}^{\slen \times d}$ using the multi-head self-attention mechanism: $Z = \zeta_T(U')$. To enforce a lower representation of the latent memory, two additional \textit{Feed Forward Layers} incorporate a bottleneck mechanism, in which the dimensions of $Z$ are concatenated and then compressed: $Z' \in \mathbb{R}^{k} = W_{enc}Z+b_{enc}$, where $W_{enc} \in \mathbb{R}^{k \times (\slen d)}$ and $b_{enc} \in \mathbb{R}^k$ are learnable parameters. Considering $k \ll \slen d$, this principle turns the Transformer into an autoencoder. A concluding \textit{Tanh Layer} after the first \textit{Feed Forward Layer} scales the compressed latent memory into the interval $(-1,1)$, which gives more control over the sampling limits when shaping the distribution of the latent space \cite{radford2015}. For the reconstruction of the time series $X$, the second \textit{Feed Forward Layer} scales the compressed latent memory $Z'$ back to the initial dimensions $Z'' \in \mathbb{R}^{\slen \times d}$ with learnable parameters $W_{dec}$ and $b_{dec}$ similar to the preceding layer, which serves as the first input for the Transformer decoder $\eta_T$. Since this model aims to optimize a reconstruction problem rather than a classification or regression problem, the second input of the Transformer decoder consists of the same input $X$, shifted to the right by $1$ time step and denoted as $\Bar{X}$. The output of the Transformer decoder corresponds to $\hat{Y} = \eta_T(\Bar{X}, Z'')$, where $\hat{Y} \in \mathbb{R}^{\slen \times v}$ is the reconstructed multivariate time series.
\newpage
\begin{figure}[!t] 
 \centering
 \includegraphics[width=.6\columnwidth]{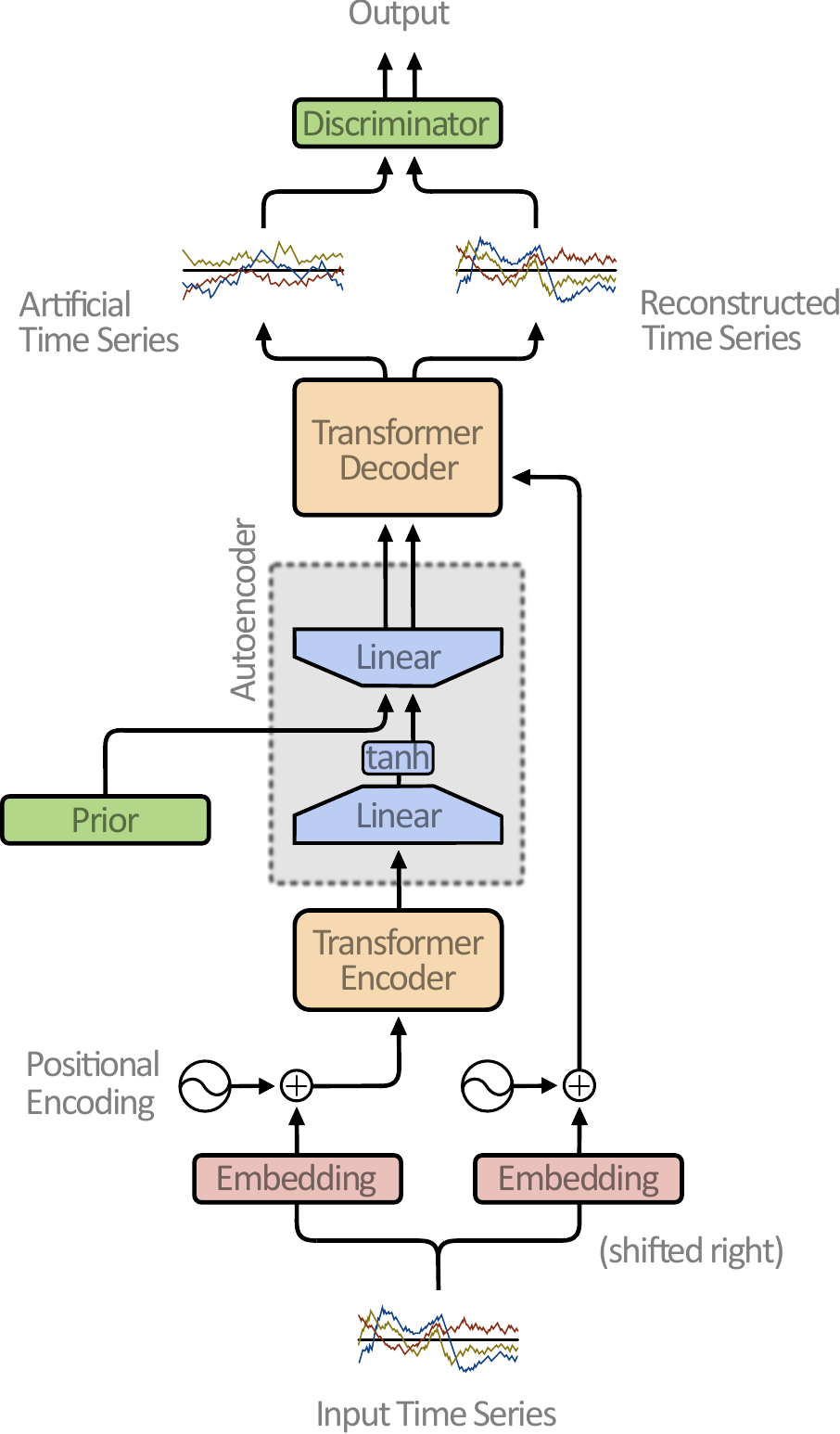} 
 \caption{The model augments the Transformer architecture with an autoencoder and an adversarial training scheme. For the Transformer decoder, the input time series is shifted to the right to serve as the output signal that is to be predicted. A memory vector is drawn from a prior distribution and is decoded to generate an artificial time series. The discriminator decides if a time series is true (from the dataset) or fake (artificially generated).}
 \label{fig:model}
\end{figure}
\noindent
The adversarial training process is incorporated into the model: The generator $G$ randomly samples a memory vector $Z$ from a uniform prior distribution with the interval $[-1,1)^k$, i.e. $Z \stackrel{iid}{\sim} U[-1,1)^k, Z \in \mathbb{R}^{k}$ and decodes the sampled memory into an artificial multivariate time series $Y \in \mathbb{R}^{\slen \times v}$ using the decoder of the autoencoder and the Transformer decoder $\eta_T$. As $\eta_T$ requires a reference time series $\bar{X}$ which does not exist for a sampled memory, the decoding procedure is done in an iterative fashion: Starting with a predefined \texttt{<SOS>}-Token, the time series is built up step-wise by appending the current output of the decoder to the time series until the maximum sequence length is reached or a predefined \texttt{<EOS>}-Token is obtained. The completed time series then corresponds to the reconstruction of the memory, i.e. a fake time series created by $G$. It is important to note that this auto-regressive, generative strategy may cause error accumulation during inference \cite{wu2020}.

The discriminator $D$ is a separate neural network. In \cite{Goodfellow2014} a theoretical proof shows that convergence of the adversarial training can be obtained if both $G$ and $D$ are given enough capacity. To avoid an imbalance of capacity and to guarantee that $D$ is not less complex than $G$, it consists of a Transformer encoder with the same model parameters as $\zeta_T$ and latent dimension $k$. The latent memory of $D$ is then passed through a \textit{Feed Forward Layer} projecting to only one neuron. A subsequent \textit{Sigmoid} function completes the architecture to make binary predictions (true/fake).

To compare the performance of this model with an existing approach, a convolutional autoencoder similar to \cite{Guidotti2020} was implemented as well. Here, the encoder $\zeta_C$ consists of $8$ 1-D \textit{Convolutional Layers} and has $u \in \{2, 4, 8, 16,\allowbreak 32, 64, 128, 256\}$ kernels of size $s \in \{21, 18, 15, 13, 11, 8,\allowbreak 5, 3\}$ per layer. A subsequent \textit{Aggregation Layer} applies a single kernel to all $256$ channels, whose output length is then compressed to the latent dimension $k$ using a \textit{Feed Forward Layer} to achieve the same autoencoder bottleneck dimensionality. The decoder $\eta_C$ has a symmetric structure with the parameters in the reverse order. Since the network operates with 1-D convolutions, all variables of the multivariate input $X \in \mathbb{R}^{\slen \times v}$ are concatenated into a single dimension $X_{concat} \in \mathbb{R}^{1 \times (\slen v)}$. Both models are implemented in Python using PyTorch. %(see \autoref{sec:Appendix}).

\subsection{Dataset and preprocessing}
\begin{figure}[!tb] 
 \centering
 \includegraphics[width=.4\columnwidth]{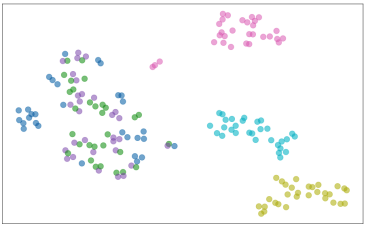} 
 \caption{t-SNE visualization of the NATOPS dataset. The colors represent the different classes (gestures).} \label{fig:tsne_natops}
\end{figure}

\begin{figure*}[h!] 
 \centering
 \includegraphics[width=\linewidth]{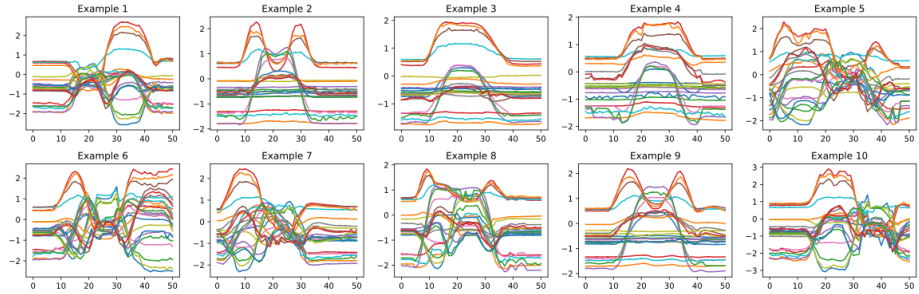} 
 \caption{Ten random examples of time series signals from the NATOPS validation set. }
 \label{fig-10timeseriesNATOPS}
\end{figure*}

In this work a subset\footnote{\href{timeseriesclassification.com/description.php?Dataset=NATOPS}{\nolinkurl{timeseriesclassification.com/description.php?Dataset=NATOPS}}} of the NATOPS dataset \cite{Song2011} is used as an exemplary small dataset, which contains body sensor recordings of gestures used as aircraft handling signals. The data is collected by sensors on the hands, elbows, wrists and thumbs which recorded the x,y,z coordinates relatively to the person. The gestures were originally recorded at $20$ FPS (with an average duration of $2.34$ seconds). In the given subset, all sequences are normalized to a sequence length of $51$ time steps. In total, the training and validation set each contain $180$ sequences with $24$ features (sensor recordings) and $6$ classes representing different gesture commands, evenly distributed over both datasets. \autoref{fig-10timeseriesNATOPS} shows ten exemplary time series. \autoref{fig:tsne_natops} shows a t-SNE representation of the whole validation set. The plot illustrates that three classes are very well separable as different modes, while the other three classes have a very similar inherent structure.
\newpage
As a preprocessing step, a feature-wise normalization is done in which the values for each feature dimension are transformed into the range $[-1,1]$ using the following equations:
\begin{equation}\label{eq3a}
 X^v_{std} = \frac{X^v-min(X^v)}{max(X^v)-min(X^v)}
\end{equation}
\begin{equation}\label{eq3b}
 X^v_{scaled} = X^v_{std} (ul - ll) + ll
\end{equation}

with the limits $ul = 1$ and $ll = -1$, where $X^v$ refers to the values of the dataset $X$ at dimension $v$. Bringing all dimensions into the same range makes sharing the model parameters more effective. The Transformer model requires a Start-Of-Sequence \texttt{<SOS>}-Token to decode and generate sequences. This is chosen to be outside of the value range and is set to $-3$. After the normalization step, the \texttt{<SOS>}-Token is prepended to each dimension for all sequences. %Das hat den Vorteil, dass die relativen Bezüge zueinander im Scaler bleiben, und die Netzwerke für alle features im gleichen Wertebereich operieren

It is important to note that time series datasets do not necessarily provide sequences of equal length. If variations in sequence lengths are observed, a maximum sequence length can be set for the Transformer autoencoder. Shorter sequences are then padded with arbitrary values and a padding mask adds a large negative value to the attention values, e.g. $-\infty$, before computing the self-attention (\autoref{eq2}). Furthermore, an \texttt{<EOS>}-Token would be needed to represent the End-Of-Sequence.

\subsection{Training schedule}
Before training, the models were initialized based on \cite{vaswani2017}. The Transformer autoencoder consists of $8$ attention heads, $6$ encoder layers and $6$ decoder layers and has a model dimension $d = 24$ which corresponds to the number of the input variables. The dimension of the hidden feedforward layer is set to $128$ and the dimension of the latent memory of the bottleneck is set to $k = 60$. The parameters are initialized using Xavier Initialization \cite{Xavier2010}. Similar to the approach of \cite{Huang2020} this initialization helps to preserve the variance of the gradients across the different layers. This choice coincides with the design of the attention mechanism in the Transformer network, which prevents the gradients of growing too large in magnitude with layer depth \cite{vaswani2017}. \\
As proposed by \cite{Makhzani2015} and \cite{yoon2019}, both the autoencoder and the adversarial networks are trained jointly on each minibatch in two phases: first, the autoencoder is updated to minimize the reconstruction error of the input. Then, the adversarial networks (i.e. the decoder of the autoencoder and the discriminator) are updated to regularize the latent space: the discriminator is trained to tell apart the true samples $X$ from the generated samples $Y$ and the generator is trained to fool the discriminator into believing the generated samples $Y$ are real. After successful training, the generator has learned a transformation that maps the imposed prior $p_{noise}(Z)$ to the data distribution $p_{data}(X)$. \\
The algorithm for the training schedule is given above. Instead of using the objective function of (\autoref{eq1}), the generator is trained to maximize $\log(D(G(Z)))$, which provides stronger gradients and thereby counteracts early vanishing \cite{Goodfellow2014}. For simplicity, the Transformer autoencoder, which consists of the encoder $\zeta_T$, the decoder $\eta_T$, and the architectural adaptions specified in \autoref{subsec:modelarchitectures}, is denoted as $\Psi$.
%\Xhline{1.5pt}
\begin{algorithm}[t!]
\small
\setlength{\abovedisplayskip}{0pt}
\setlength{\belowdisplayskip}{0pt}
\caption{Simultaneous minibatch SGD training of the autoencoder and the GAN structure.}
\For{number of epochs}{
\For{number of batches}{
\begin{itemize}[leftmargin=*]
\item Sample minibatch of $b$ samples $\{X^{(1)},...,X^{(b)}\}$\\ from data distribution $p_{data}(X)$
\item Update the autoencoder by its stochastic gradient:
\begin{fleqn}\begin{equation*}
~~~~\nabla_{\theta_\Psi} \, \frac{1}{b}\sum_{i=1}^{b}\Vert X^{(i)}-\Psi(X^{(i)})\Vert_2
\end{equation*}\end{fleqn}
\item Sample minibatch of $b$ fake samples $\{Y^{(1)},...,Y^{(b)}\}$ \\using the generator $Y=G(Z)$ and $Z \sim p_{noise}(Z)$
\item Update the discriminator by its stochastic gradient:
\begin{fleqn}\begin{equation*}
~~~~\nabla_{\theta_D} \, \frac{1}{b}\sum_{i=1}^{b}[\log(D(X^{(i)}))+\log(1-D(Y^{(i)}))]
\end{equation*}\end{fleqn}
\item Update the generator by its stochastic gradient:
\begin{fleqn}\begin{equation*}
~~~~\nabla_{\theta_G} \, \frac{1}{b}\sum_{i=1}^{b}[\log(D(Y^{(i)}))]
\end{equation*}\end{fleqn}
\end{itemize}
} % for
} % for
%\noindent \textbf{for} number of epochs \textbf{do} \\
%\indent \textbf{for} number of batches \textbf{do} \\
%\indent \indent $\bullet$ Sample minibatch of $b$ samples $\{X^{(1)},...,X^{(b)}\}$ \\
%\indent \indent \hspace{5pt} from data distribution $p_{data}(X)$\\
%\indent \indent $\bullet$ Update the autoencoder by its stochastic gradient:\\
%\vspace{-15pt}
%\begin{align*}
%\nabla_{\theta_\Psi} \hspace{3pt} \frac{1}%{b}\sum_{i=1}^{b}\Vert X^{(i)}-\Psi(X^{(i)})\Vert_2
%\end{align*}
%\indent \indent $\bullet$ Sample minibatch of $b$ fake samples $\{Y^{(1)},...,Y^{(b)}\}$ \\
%\indent \indent \hspace{5pt} using the generator $Y=G(Z)$ and $Z \sim p_{noise}(Z)$ \\
%\indent \indent $\bullet$ Update the discriminator by %its stochastic gradient:\\
%\vspace{-15pt}
%\begin{align*}
%\nabla_{\theta_D} \hspace{3pt} \frac{1}{b}\sum_{i=1}^{b}%[log(D(X^{(i)}))+log(1-D(Y^{(i)}))]
%\end{align*}
%\indent \indent $\bullet$ Update the generator by its stochastic gradient:\\
%\vspace{-15pt}
%\begin{align*}
%\nabla_{\theta_G} \hspace{3pt} \frac{1}{b}\sum_{i=1}^{b}[log(D(Y^{(i)}))]
%\end{align*}
%\indent \textbf{end for}\\
%\textbf{end for}\\
\vspace{0.2cm}
\footnotesize{Adam optimizer \cite{Kingma2014} is used to incorporate the gradient-based updates into a learning rule.}
%\midrule
%\noindent
\end{algorithm}
%\toprule
%\noindent
Given the NATOPS dataset, the Transformer autoencoder was trained for $2\,000$ epochs, with a batch size equals to $32$ and a learning rate of $10^{-4}$. The training was pursued  at the Training Center for Machine Learning (TCML) Cluster in Tübingen (Grant number 01IS17054). The convolutional autoencoder was trained using the same procedure. However, all weights were instead initialized from a zero-centered Normal distribution with a standard deviation of $0.02$ \cite{radford2015} and a learning rate of $0.001$ was used.%with the xxxxxxxxxxxxxxxxx cluster (grant number xxxxxxxxxx) the University of xxxxxxxxx in xxxxxxxx.

A variant of the above algorithm is yielded by incorporating the Wasserstein GAN scheme \cite{Arjovsky2017_wsGAN} to stabilize GAN training. Here, both the discriminator and the generator are optimized using an adapted loss function (eq. \eqref{eq7}), the discriminator uses a linear activation instead of sigmoid, is trained $5$ times more than the generator in each iteration and its weights are constrained to a range of $[-0.1,0.1]$ after each update. Furthermore RMSprop is used as learning rule with a small learning rate of $5\cdot10^{-5}$ and without momentum.
\begin{equation}
\label{eq7}
\max\mathbb{E}_{x\sim p_{data}}D(x)-\mathbb{E}_{z\sim p_{noise}}D(G(z))
\end{equation}

\subsection{Evaluation metrics}
The characteristic of the latent space and the generation of artificial multivariate time series can be evaluated qualitatively and quantitatively \cite{Brophy2021}. However, unlike image-based GANs, where the Inception Score \cite{salimans2016} or the Fréchet Inception Distance (FID) \cite{Heusel2017} are established metrics, there are no standards set for time series data, especially not for multivariate time series. Borji in \cite{Borji2018} gives an overview of different evaluation metrics, which however are mainly focused on image generation and for the most part lack possibilities to adapt to (multivariate) time series data.

In this work, t-SNE visualizations \cite{vandermaaten2008} are used to qualitatively compare the distribution of artificially generated signals with the underlying distribution of the dataset. To quantitatively measure the similarity of both distributions Dynamic Time Warping (DTW) is applied, which can be generalized to a dependent, multi-dimensional form proposed by \cite{shokoohi2017}. More specifically, we use the DTW distance that is the accumulated ``warping costs'' to optimally match two sequences. Note that DTW is more robust against time lags than other similarity or distance measures like the correlation coefficient or the euclidean distance. For each generated multivariate time series, DTW is used to determine the shortest distance to a signal from the validation set. The average DTW is then calculated as the mean of all shortest distances. The smaller the distance, the higher the similarity. To additionally quantitatively assess and compare the diversity of the generated time series, the multivariate Entropy is calculated based on \cite{Bahrpeyma2021}. Thereby the total Entropy $H$ is determined by the sum of the Entropy per dimension $H_E$, normalized by the maximum Entropy $H_{max}$ as shown by  \autoref{eq4} -- \autoref{eq6} in the following:
\begin{equation}
\label{eq4}
H_{max}(X) = -\sum_{i=1}^{S} \frac{1}{|S|} \log \frac{1}{|S|} = \log |S|
\end{equation}
\begin{equation}
\label{eq5}
H_{E}(X) = -\frac{1}{H_{max}} \sum_{i=1}^{S} p(x_i) \log p(x_i) 
\end{equation}
\begin{equation}
\label{eq6}
H = \frac{1}{k} \sum_{i=1}^{k} H_{E}(X^i) 
\end{equation}
where $k$ is the number of dimensions and $S$ is a set of probabilities. To define $S$ domain knowledge about the dataset is required: each dimension is categorized into value ranges and by counting the occurrence of time points within these ranges, the probability for each category is determined. Four categories are determined for the given dataset, where $X^i_t$ refers to all artificial signals at dimension $i$ and time step $t$: $p_1: X^i_t \geq 1, p_2: 1 > X^i_t \geq 0, p_3: 0 > X^i_t \geq -1, p_4: X^i_t < -1$. The higher the score, the higher the diversity. For each model $50$ artificial time series were generated to calculate the scores of the metrics.

\section{Results}
Both the Transformer autoencoder and the convolutional autoencoder were train-ed with and without the GAN training scheme. Training without GAN only involved minimizing the reconstruction error without any regularization of the latent space. Additionally, both models were trained using the Wasserstein GAN approach. In this section the models are abbreviated the following: TAE = Transformer autoencoder, TAE-GAN = Transformer autoencoder with GAN scheme, TAE-WGAN = Transformer autoencoder with Wasserstein GAN scheme. The convolutional autoencoder is abbreviated in the same manner using the basis CAE. \autoref{fig:tsne} shows the t-SNE visualizations of all models. \autoref{tab:results} contains the results of all models regarding similarity (Avg. DTW), diversity (Entropy) and the reconstruction ability of the original signals (Test Error). 

\subsubsection{TAE:} 
\begin{figure*}[!tb] 
 \centering
 \includegraphics[width=\linewidth]{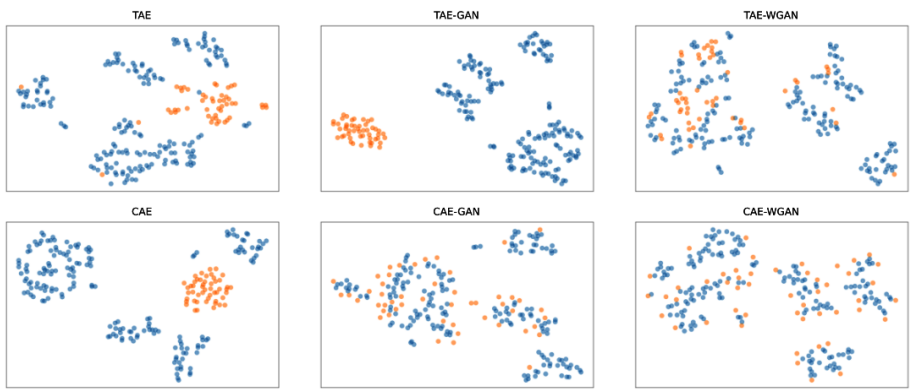} 
 \caption{t-SNE representation of all models. The blue dots refer to the true time series from the validation dataset, while the orange dots refer to artificially generated time series from the models. Models without GAN regularization (TAE, CAE) did not learn the representation. TAE-GAN seems to have suffered from mode dropping and mode collapse.}
 \label{fig:tsne}
\end{figure*}

The model shows a higher diversity (Entropy) compared to the TAE trained with GAN. The t-SNE representation in \autoref{fig:tsne} illustrates this variance, but the generated time series are hardly represented within the modes of the dataset, as the latent space was not shaped any further. The reconstruction error is smaller compared to the TAE trained with GAN. Apparently, the GAN training counteracted decoding accuracy as a regulatory constraint. Although TAE achieved the lowest reconstruction error, it is important to note that the Transformer model generally suffers from error accumulation when generating artificial time series signals, because the generation process is done in an iterative fashion.

\begin{table}[b!]
\centering
\caption{Average DTW, multivariate Entropy and reconstruction error after training.}
\label{tab:results}
 {\small
 \begin{tabularx}{\linewidth}{CCCC}
  \toprule
  \textbf{Model Type} & \textbf{Avg. DTW} & \textbf{Entropy} & \textbf{Test Error} 
  \\
  \midrule
  %\vspace{1pt}
  TAE & $28.273$ & $0.544$ & $\textbf{0.018}$ \\
  TAE-GAN & $36.918$ & $0.427$ & $0.283$ \\
  TAE-WGAN & $\textbf{19.919}$  & $0.394$ & $0.019$\\
  \midrule
  CAE & $36.524$  & $0.284$ & $0.082$ \\
  CAE-GAN & $35.164$ & $0.589$ & $0.047$ \\
  CAE-WGAN & $37.949$ & $\textbf{0.855}$ & $0.024$ \\
  \bottomrule
\end{tabularx}}
\end{table}

\subsubsection{TAE-GAN:} Both the diversity (Entropy) and the similarity (DTW) are worse compared to TAE. This is also suggested by the t-SNE representation, since only one mode is learned that does not overlap with the original distribution. This is a typical example of 1. Mode dropping and 2. Mode collapse, where 1. mass is put outside of the support of the underlying distribution of the dataset and 2. only a single type of output is learned and other modes are disregarded \cite{Lucas2020}. 

\subsubsection{TAE-WGAN:} 
\begin{figure*}[h!] 
 \centering
 \includegraphics[width=1\linewidth]{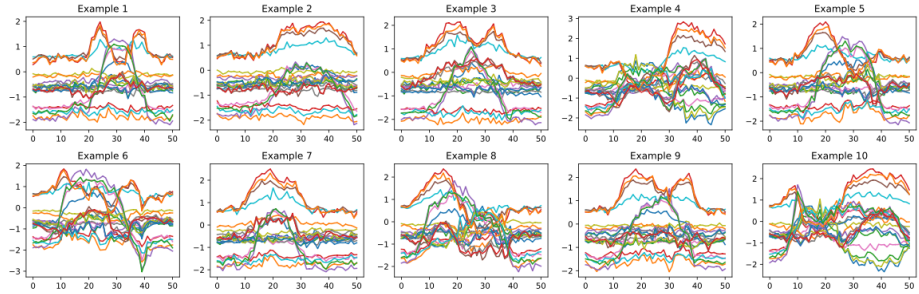} 
 \caption{Ten different time series samples generated by TAE-WGAN. They show similar patterns to the validation dataset, however contain small fluctuations.}
 \label{fig-10timeseriesTWGAN}
\end{figure*}

With a score of $19.919$, the similarity (DTW) to the dataset is the highest compared to all other models. However, the diversity (Entropy) has deteriorated compared to TAE and TAE-GAN. The t-SNE representation in \autoref{fig:tsne} actually suggests a higher score due to the higher dispersion, but it also shows that the three well-separated modes are not sufficiently represented. The reconstruction ability was hardly influenced by the GAN regularization compared to TAE (similar Test Error) and is smaller compared to the convolutional models. The generated time series in \autoref{fig-10timeseriesTWGAN} have similar patterns to those from the dataset and indeed show some diversity. However, they contain fluctuations where straight lines or smooth curves were expected.

\subsubsection{CAE:} This model shows low similarity (DTW) and low diversity (Entropy) compared to the other models. The low diversity is further illustrated by the distribution in the t-SNE representation in \autoref{fig:tsne}. Similar to TAE-GAN, the distribution lies exclusively within a new mode; there is no overlap with the modes of the dataset as the latent space was not regularized any further. 

\subsubsection{CAE-GAN:} The Entropy score shows a high diversity compared to the other models, also confirmed by the distribution in the t-SNE plot (\autoref{fig:tsne}). Here, the artificial time series lie within the different modes of the dataset, although not sufficiently distributed. The similarity score (DTW) is still low in an overall comparison, however slightly decreased compared to CAE. 

\subsubsection{CAE-WGAN:} With an Entropy of $0.855$ this model shows the highest diversity, which is also confirmed by the distribution in the t-SNE plot (\autoref{fig:tsne}). The generated time series are well balanced among the different modes of the dataset. Nevertheless, the similarity (DTW) remained low and even worsened compared to CAE and CAE-GAN. 

\section{Conclusion}
In this work, a Transformer-based model is proposed to generate multivariate time series signals within the support of the underlying distribution of an existing dataset. The Transformer architecture was augmented with a bottleneck to act as an autoencoder and was optimized using a conventional GAN and the Wasserstein GAN approach. Additionally, the model was compared to an existing approach that uses a convolutional autoencoder for local neighborhood generation of time series signals. The results show that the Transformer model was more suitable for stable GAN training. Using the Wasserstein GAN approach, the model outperformed the other models in terms of similarity to the dataset, however not all modes were sufficiently represented. The diversity is estimated to be rather low based on the multivariate Entropy, while the t-SNE representation suggests a higher value. This discrepancy underscores the need for profound research on appropriate GAN evaluation metrics for multivariate time series. It is however important to note that distances in t-SNE representations are not necessarily comparable due to nonlinear transformations. Using the conventional GAN training scheme, longer training could have resulted in convergence and higher similarity, however, harbors the risk of overfitting to the training data and thereby leading to phenomena like mode dropping. Artifacts such as small fluctuations in the time series unfortunately could not be diminished. Regularizing GANs is an ongoing research problem, and the behavior of GANs needs to be further understood.

In summary, our proposed Transformer-based Wasserstein GAN archictecture is a promising candidate for better representation learning of multivariate time series and could have an impact to related domains such as data augmentation in various fields involving multivariate time series \cite{MoralesEsteban2010,Moustra2011,He2015,Bui2017,Balaha2022,Ding2020,Severin2020}
or generating counterfactual explanations \cite{lang2023generating}.

\section{Code availability}
All Python scripts regarding the model architectures, training, loading the dataset, and evaluation are available at \\ \href{https://github.com/lscharwaechter/TransformerGAN}{https://github.com/lscharwaechter/TransformerGAN}.

\bibliographystyle{splncs04}
\bibliography{References}

\begin{thebibliography}{10}
\providecommand{\url}[1]{\texttt{#1}}
\providecommand{\urlprefix}{URL }
\providecommand{\doi}[1]{https://doi.org/#1}

\bibitem{Arjovsky2017_wsGAN}
Arjovsky, M., Chintala, S., Bottou, L.: Wasserstein gan. arXiv:1701.07875  (2017)

\bibitem{Assaf2019}
Assaf, R., Schumann, A.: Explainable deep neural networks for multivariate time series predictions. In: Proceedings of the Twenty-Eighth International Joint Conference on Artificial Intelligence. International Joint Conferences on Artificial Intelligence Organization (aug 2019)

\bibitem{Ates2021}
Ates, E., Aksar, B., Leung, V.J., Coskun, A.K.: Counterfactual explanations for multivariate time series. In: International Conference on Applied Artificial Intelligence ({ICAPAI}). {IEEE} ((2021))

\bibitem{Bahrpeyma2021}
Bahrpeyma, F., Roantree, M., Cappellari, P., Scriney, M., McCarren, A.: A methodology for validating diversity in synthetic time series generation. {MethodsX}  \textbf{8},  101459 (2021)

\bibitem{Balaha2022}
Balaha, H.M., Shaban, A.O., El-Gendy, E.M., Saafan, M.M.: A multi-variate heart disease optimization and recognition framework. Neural Computing and Applications  \textbf{34}(18),  15907--15944 (May 2022)

\bibitem{Borji2018}
Borji, A.: Pros and cons of gan evaluation measures. arXiv:1802.03446  (2018)

\bibitem{Brophy2021}
Brophy, E., Wang, Z., She, Q., Ward, T.: Generative adversarial networks in time series: A survey and taxonomy. arXiv:2107.11098  (2021)

\bibitem{Bui2017}
Bui, C., Pham, N., Vo, A., Tran, A., Nguyen, A., Le, T.: Time Series Forecasting for Healthcare Diagnosis and Prognostics with the Focus on Cardiovascular Diseases, pp. 809--818. Springer Singapore (Sep 2017)

\bibitem{Ding2020}
Ding, H., Guo, L., Zhao, C., Wang, F., Wang, G., Jiang, Z., Xi, W., Zhao, J.: Rfnet: Automatic gesture recognition and human identification using time series rfid signals. Mobile Networks and Applications  \textbf{25}(6),  2240--2253 (Nov 2020)

\bibitem{Xavier2010}
Glorot, X., Bengio, Y.: Understanding the difficulty of training deep feedforward neural networks. In: Teh, Y.W., Titterington, M. (eds.) Proceedings of the Thirteenth International Conference on Artificial Intelligence and Statistics. Proceedings of Machine Learning Research, vol.~9, pp. 249--256. PMLR, Chia Laguna Resort, Sardinia, Italy (13--15 May 2010)

\bibitem{Goodfellow2014}
Goodfellow, I.J., Pouget-Abadie, J., Mirza, M., Xu, B., Warde-Farley, D., Ozair, S., Courville, A., Bengio, Y.: Generative adversarial networks (2014)

\bibitem{Guidotti2020}
Guidotti, R., Monreale, A., Spinnato, F., Pedreschi, D., Giannotti, F.: Explaining any time series classifier. In: Second International Conference on Cognitive Machine Intelligence ({CogMI}). {IEEE} ((2020))

\bibitem{He2015}
He, G., Duan, Y., Peng, R., Jing, X., Qian, T., Wang, L.: Early classification on multivariate time series. Neurocomputing  \textbf{149},  777--787 (Feb 2015)

\bibitem{Heusel2017}
Heusel, M., Ramsauer, H., Unterthiner, T., Nessler, B., Hochreiter, S.: Gans trained by a two time-scale update rule converge to a local nash equilibrium. arXiv:1706.08500  (2017)

\bibitem{Huang2020}
Huang, X.S., Pérez, F., Ba, J., Volkovs, M.: Improving transformer optimization through better initialization. In: Proceedings of the 37th International Conference on Machine Learning. ICML'20, JMLR.org (2020)

\bibitem{Khan2022}
Khan, S., Naseer, M., Hayat, M., Zamir, S.W., Khan, F.S., Shah, M.: Transformers in vision: A survey. ACM Comput. Surv.  \textbf{54}(10s) (sep 2022)

\bibitem{Kingma2014}
Kingma, D.P., Ba, J.: Adam: A method for stochastic optimization. arXiv:1412.6980  (2014)

\bibitem{lang2023generating}
Lang, J., Giese, M.A., Ilg, W., Otte, S.: Generating sparse counterfactual explanations for multivariate time series. In: International Conference on Artificial Neural Networks (ICANN). pp. 180--193. Springer (2023)

\bibitem{ledig2017}
Ledig, C., Theis, L., Huszár, F., Caballero, J., Cunningham, A., Acosta, A., Aitken, A., Tejani, A., Totz, J., Wang, Z., et~al.: Photo-realistic single image super-resolution using a generative adversarial network. In: Proceedings of the IEEE conference on computer vision and pattern recognition. pp. 4681--4690 (2017)

\bibitem{li2022}
Li, X., Metsis, V., Wang, H., Ngu, A.H.H.: Tts-gan: A transformer-based time-series generative adversarial network. In: Artificial Intelligence in Medicine: 20th International Conference on Artificial Intelligence in Medicine, AIME 2022, Halifax, NS, Canada, June 14--17, 2022, Proceedings. pp. 133--143. Springer (2022)

\bibitem{Lin2022}
Lin, T., Wang, Y., Liu, X., Qiu, X.: A survey of transformers. AI Open  \textbf{3},  111--132 (2022)

\bibitem{Lucas2020}
Lucas, T.: Deep generative models: over-generalisation and mode-dropping. Ph.D. thesis, Artificial Intelligence [cs.AI]. Université Grenoble Alpes (2020)

\bibitem{vandermaaten2008}
Van~der Maaten, L., Hinton, G.: Visualizing data using t-sne. Journal of machine learning research  \textbf{9}(11) (2008)

\bibitem{Makhzani2015}
Makhzani, A., Shlens, J., Jaitly, N., Goodfellow, I.J.: Adversarial autoencoders. arXiv:1511.05644  (2015)

\bibitem{MoralesEsteban2010}
Morales-Esteban, A., Martínez-Álvarez, F., Troncoso, A., Justo, J., Rubio-Escudero, C.: Pattern recognition to forecast seismic time series. Expert Systems with Applications  \textbf{37}(12),  8333--8342 (Dec 2010)

\bibitem{Moustra2011}
Moustra, M., Avraamides, M., Christodoulou, C.: Artificial neural networks for earthquake prediction using time series magnitude data or seismic electric signals. Expert Systems with Applications  \textbf{38}(12),  15032--15039 (Nov 2011)

\bibitem{radford2015}
Radford, A., Metz, L., Chintala, S.: Unsupervised representation learning with deep convolutional generative adversarial networks (2015)

\bibitem{salimans2016}
Salimans, T., Goodfellow, I., Zaremba, W., Cheung, V., Radford, A., Chen, X.: Improved techniques for training gans. Advances in neural information processing systems  \textbf{29} (2016)

\bibitem{Severin2020}
Severin, I.C.: Time series feature extraction for head gesture recognition: Considerations toward hci applications. In: 2020 24th International Conference on System Theory, Control and Computing (ICSTCC). IEEE (Oct 2020)

\bibitem{shokoohi2017}
Shokoohi-Yekta, M., Hu, B., Jin, H., Wang, J., Keogh, E.: Generalizing dtw to the multi-dimensional case requires an adaptive approach. Data mining and knowledge discovery  \textbf{31},  1--31 (2017)

\bibitem{Song2011}
Song, Y., Demirdjian, D., Davis, R.: Tracking body and hands for gesture recognition: Natops aircraft handling signals database. In: 2011 IEEE International Conference on Automatic Face \& Gesture Recognition (FG). pp. 500--506 (2011)

\bibitem{Tran2018dist}
Tran, N.T., Bui, T.A., Cheung, N.M.: Dist-gan: An improved gan using distance constraints. In: Computer Vision -- ECCV 2018. pp. 387--401. Springer International Publishing, Cham (2018)

\bibitem{vaswani2017}
Vaswani, A., Shazeer, N., Parmar, N., Uszkoreit, J., Jones, L., Gomez, A.N., Kaiser, L., Polosukhin, I.: Attention is all you need. arXiv:1706.03762  (2017)

\bibitem{wu2020}
Wu, S., Xiao, X., Ding, Q., Zhao, P., Wei, Y., Huang, J.: Adversarial sparse transformer for time series forecasting. Advances in neural information processing systems  \textbf{33},  17105--17115 (2020)

\bibitem{yoon2019}
Yoon, J., Jarrett, D., Van~der Schaar, M.: Time-series generative adversarial networks. Advances in neural information processing systems  \textbf{32} (2019)

\bibitem{Zerveas2021}
Zerveas, G., Jayaraman, S., Patel, D., Bhamidipaty, A., Eickhoff, C.: A transformer-based framework for multivariate time series representation learning. pp. 2114--2124. Association for Computing Machinery, New York, NY, USA (2021)

\bibitem{zhang2017}
Zhang, H., Xu, T., Li, H., Zhang, S., Wang, X., Huang, X., Metaxas, D.N.: Stackgan: Text to photo-realistic image synthesis with stacked generative adversarial networks. In: Proceedings of the IEEE international conference on computer vision. pp. 5907--5915 (2017)

\bibitem{zhang2022}
Zhang, J., Dai, Q.: Latent adversarial regularized autoencoder for high-dimensional probabilistic time series prediction. Neural Networks  \textbf{155},  383--397 (2022)

\end{thebibliography}

\end{document}